\documentclass{article}

\usepackage{times}
\usepackage{graphicx} 

\usepackage{natbib}

\usepackage{algorithm}
\usepackage{algorithmic}
\usepackage{color}

\usepackage{amsmath}
\usepackage{amsfonts}

\usepackage{hyperref}

\usepackage[accepted]{icml2016}

\begin{document} 

\twocolumn[
\icmltitle{Residual Convolutional CTC Networks for Automatic Speech Recognition}


\icmlauthor{Yisen Wang$^{*\dag\ddag}$}{wangys14@mails.tsinghua.edu.cn}
\icmlauthor{Xuejiao Deng$^{*\dag}$}{sophiadeng@tencent.com}
\icmlauthor{Songbai Pu$^{\dag}$}{johnsonpu@tencent.com}
\icmlauthor{Zhiheng Huang$^{\dag}$}{zhihhuang@tencent.com}
\icmladdress{$^\dag$Tencent AI Lab, Tencent, Shenzhen, China}
\icmladdress{$^\ddag$Department of Computer Science and Technology, Tsinghua University, Beijing, China}





\icmlkeywords{Residual CNN, CTC, System Combination, ASR}

\vskip 0.3in
]


\begin{abstract} 
Deep learning approaches have been widely used in Automatic Speech Recognition (ASR) and they have achieved a significant accuracy improvement. Especially, Convolutional Neural Networks (CNNs) have been revisited in ASR recently. However, most CNNs used in existing work have less than 10 layers which may not be deep enough to capture all human speech signal information. In this paper, we propose a novel deep and wide CNN architecture denoted as RCNN-CTC, which has residual connections and Connectionist Temporal Classification (CTC) loss function. RCNN-CTC is an end-to-end system which can exploit temporal and spectral structures of speech signals simultaneously. Furthermore, we introduce a CTC-based system combination, which is different from the conventional frame-wise senone-based one. The basic subsystems adopted in the combination are different types and thus mutually complementary to each other. Experimental results show that our proposed single system RCNN-CTC can achieve the lowest word error rate (WER) on WSJ and Tencent Chat data sets, compared to several widely used neural network systems in ASR. In addition, the proposed system combination can offer a further error reduction on these two data sets, resulting in relative WER reductions of $14.91\%$ and $6.52\%$ on WSJ dev93 and Tencent Chat data sets respectively.


\end{abstract} 

\section{Introduction}
Automatic Speech Recognition (ASR) is designed to automatically transcribe human speech into text. In the past several years, deep learning \cite{Yu:2014:ASR:2695502} has been successfully applied in ASR to boost the recognition accuracy. Very recently, CNN becomes an attractive model in ASR, which transforms speech signals into feature maps as used in computer vision \cite{lecun1998convolutional}. Compared to other deep learning architectures, CNN has several advantages: 1) CNN is suited to exploit local correlations of human speech signals in both time and frequency dimensions. 2) CNN has the capacity to exploit translational invariance in signals. 

Most of previous applications of CNN in ASR only used a few convolutional layers. One typical architecture usually contains several convolutional layers, followed by a number of recurrent layers and fully-connected feedforward layers. These CNN structures are often less than 10 layers\footnote{One exception is LACE \cite{yu2016deep} which has about 20 layers, but it does not utilize CTC as proposed in this paper.}, which may not be deep enough to capture all the information of human speech signals, especially for long sequences. As a result, their WERs may be adversely affected. Also, the convergence speed is too slow for training this type of architecture for acoustic models in practice.  

Traditional acoustic model training is based on frame-wise cross entropy loss (CE), which requires pre-generated and aligned frame labels by hidden Markov model/Gaussian mixture model (HMM/GMM) paradigm. To simplify this process, \citeauthor{graves2006connectionist} (\citeyear{graves2006connectionist}) introduced CTC objective function to infer speech-label alignments automatically without any intermediate process, leading to an end-to-end system for ASR. CTC technique has shown promising results in Deep Speech \cite{hannun2014deep,amodei2015deep} and EESEN \cite{miao2015eesen}.




Motivated by the above observations, a residual convolutional neural networks architecture along with CTC loss system, denoted as RCNN-CTC, is proposed in this paper to boost the performance of ASR. RCNN-CTC has the following three advantages: 1) It is a CNN-based system which operates on both time and frequency dimensions. RCNN-CTC can model temporal as well as spectral local correlations and gain translational invariance in speech signals. 2) Its network architecture can be very deep (more than 40 layers) to obtain more expressive power and better generalization capacity through residual connections between layers, as inspired by Residual Networks (ResNets) \cite{he2016deep}. 3) RCNN-CTC can also be trained in an end-to-end manner thanks to the CTC loss. In addition to the proposed RCNN-CTC, we propose a CTC-based system combination to further enhance the recognition accuracy. The proposed combination is different from the conventional frame-wise senone-based one due to the fact that the former produces peak phone/label distribution while the latter produces frame-wise senone distribution. The basic subsystems adopted in our combination are RCNN-CTC, Bidirectional Long Short Term Memory (BLSTM) \cite{sak2014long} and Convolutional Long short term memory Deep Neural Network (CLDNN) \cite{sainath2015learning}. They have heterogeneous structures and are mutually complementary in producing transcriptions (see Section 4). Note that the CTC-based system combination may be difficult as the output of each basic subsystem is not frame-aligned and the scores are not well calibrated, thus the results cannot be simply averaged. We implement a series of procedures of time normalization, alignment and voting to address the above issue. 


In summary, our contributions in this paper are threefolds: 1) We propose a residual convolutional neural networks architecture paired with CTC loss (RCNN-CTC) for ASR task. Such a deep and wide network has not been applied to ASR before in our knowledge; 2) We propose a novel CTC-based system combination, which can obtain significant reduction on WER in our experiments; 3) Empirically, our proposed single system RCNN-CTC can achieve lower WERs compared with other widely used neural network ASR systems on WSJ and Tencent Chat data sets. In addition, the proposed system combination can further reduce the WERs on these two data sets.




\section{Related Work}
In the last few years, Recurrent Neural Networks (RNNs) have been widely used for sequential modeling due to its capability of modeling long history \cite{mikolov2010recurrent}. As a sequential task, Long Short Term Memory (LSTM) \cite{hochreiter1997long} and Bidirectional LSTM (BLSTM) \cite{graves2005framewise} have also been successfully applied to ASR, and they addressed the drawbacks of RNN, such as the gradient vanishing problem. However, a disadvantage of LSTM is that it needs to store multiple gating neural responses at each time-step and unfold the time steps during training and test stages, which results in a computational bottleneck for long sequences, \textit{i.e.}, thousands of frames in ASR. CNN was introduced into ASR to alleviate the computational problem. In early work, only a few CNN layers were typically used. For example,  \citeauthor{abdel2014convolutional} (\citeyear{abdel2014convolutional}) used one convolutional layer, one pooling layer and a few fully-connected layers. \citeauthor{amodei2015deep} (\citeyear{amodei2015deep}) used three convolutional layers as the feature preprocessing layers.  \citeauthor{palaz2015analysis} (\citeyear{palaz2015analysis}) showed that CNN-based speech recognition which uses raw speech as input can be more robust. To the end, deep CNN (about 10 convolutional layers) showed great performance in noisy speech recognition \cite{qian2016very,sercu2016advances}. 

Recently, ResNet \cite{he2016deep} has been shown to achieve compelling convergence and high accuracy in computer vision, which attributes to its identity mapping as the skip connection in residual blocks. Successful attempts along this line in ASR have also been reported very recently. \citeauthor{zhang2016very} (\citeyear{zhang2016very}) proposed a deep convolutional network with batch normalization (BN), residual connections and convolutional LSTM structure. Convolutional LSTM uses convolutions to replace the inner products within LSTM units. Residual connections are used to train very deep network, and BN normalizes each layer's inputs to reduce internal covariance shift. The above techniques are employed to add more computation depth to the model while reducing the number of parameters at the same time. Another network architecture was proposed in \cite{zhang2016deep}, \textit{i.e.}, deep recurrent convolutional network with deep residual learning. They implemented several recurrent layers at the bottom, followed by deep full convolutional layers with $3 \times 3$ filters (but no pooling layer). Besides, they built four residual blocks among the CNN layers, with each residual block containing layers with the same number of feature maps to avoid extra parameters. Residual LSTM architecture was proposed in \cite{kim2017residual}. In addition to the inherent shortcut paths between LSTM memory cells, they employed additional spatial shortcut paths between layer outputs. They showed that the residual LSTM architecture provided a large gain from increasing depth. However, these models still suffer from the computational bottleneck, due to the components of LSTM in their network architectures.  








\citeauthor{yu2016deep} (\citeyear{yu2016deep}) proposed another deep CNN with layer-wise context expansion and location-based attention architecture (LACE). The layer-wise context expansion and location-based attention mechanism are implemented by element-wise matrix product and convolution operations without max-pooling or average-pooling. Moreover, they employed four residual blocks, each having an identical structure, which is similar to ResNet. It is worth pointing out that they did not employ CTC loss. Consequently, LACE depends on the tedious label alignment process and cannot facilitate an end-to-end training framework. 






\section{Residual Convolutional CTC Networks}
As stated above, CNN and CTC both own excellent characteristics for ASR task, but the combination of these two components is not fully explored. In this paper, we propose a novel residual convolutional CTC networks architecture, namely RCNN-CTC, which is very deep (more than 40 layers) to get full value of CNN, residual connections and CTC. 


\subsection{Residual CNN}
\begin{figure}[!t]
\vskip 0.2in
    \centering
    \includegraphics[scale=0.6]{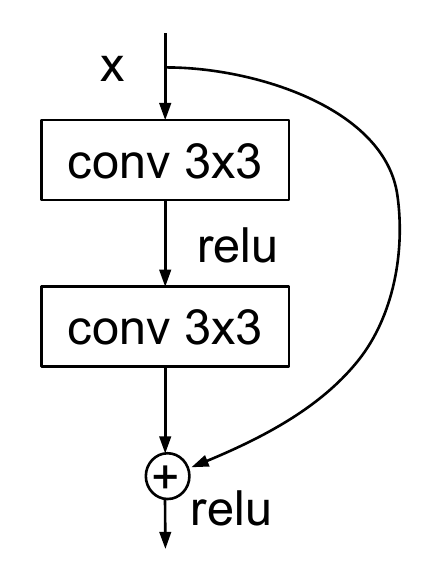}
    \caption{The structure of residual block (ResBlock). Generally, it contains two convolutional layers with small $3\times3$ filter. Each layer has an identical structure, and the skip connection is the identity mapping of $\textbf{x}$.}
    \label{ResBlock}
\end{figure}

Generally speaking, deep CNNs can improve generalization and outperform shallow networks. However, they tend to be more difficult to train and slower to converge. Residual Networks (ResNets) \cite{he2016deep} have been proposed recently to ease the training of very deep CNNs. ResNet is composed of a number of stacked residual blocks, and each block contains direct links between the lower layer outputs and the higher layer inputs. The residual block (described in Figure \ref{ResBlock}) is defined as:
\begin{equation}
\textbf{y} = \mathcal{F}(\textbf{x}, \textbf{W}_i) + \textbf{x},
\end{equation}
where $\textbf{x}$ and $\textbf{y}$ are the input and output of the layers considered, and $\mathcal{F}$ is the stacked nonlinear layers mapping function. Note that identity shortcut connections of $\textbf{x}$ do not add extra parameters and computational complexity. With the presence of residual connections, ResNet can improve the convergence speed in training. ResNet can also enjoy accuracy gains from greatly increased depth, producing results substantially better than previous networks. 

Recently, \citeauthor{zagoruyko2016wide} (\citeyear{zagoruyko2016wide}) showed that wide residual networks (WRNs) are superior over the commonly used narrow and very deep counterparts (original ResNets), which widens the convolutional layers by adding more feature maps in each residual block. Note that more feature maps mean more computation. In order to get a trade-off between performance and computational complexity, we adopt the network architecture with $width = 2$, \textit{i.e.}, our network is $2$ times wider of original ResNets architecture. The details of the proposed RCNN-CTC network architecture is shown in Table \ref{RCNN-CTC-arctec}. In particular, we use a large $41 \times 11$ filter with $32$ feature maps and $width = 1$ as $conv1$, followed by 4 groups (each with size $N$, $width =2$) of residual blocks defined in Figure \ref{ResBlock}, namely $ResBlock1$, $ResBlock2$, $ResBlock3$ and $ResBlock4$ ($N = 5$ and $2$ for Tencent Chat and WSJ data respectively, due to the fact that the former data is larger than the latter). 

\newcommand{\blocka}[2]{
  \(\left[
      \begin{array}{c}
        \text{3$\times$3, #1}\\[-.1em]
        \text{3$\times$3, #1}
      \end{array}
    \right]\)$\times$#2
}
\newcommand{\blockb}[2]{
  \(\left[
      \begin{array}{c}
        \text{3$\times$3, #1}\\[-.1em]
        \text{1$\times$1, #1}\\[-.1em]
        \text{3$\times$3, #1}
      \end{array}
    \right]\)$\times$#2
}
\newcommand{\convsize}[1]{#1$\times$#1}
\newcommand{\convname}[1]{#1}
\def\cellheight{0.34cm}
\begin{table}[!t]
  \caption{Network architecture of RCNN-CTC. It comprises of 4 groups of residual blocks with small $3\times3$ filters. The size of each block is $N$ and the width is 2. For different convolutional layers, we set different strides to reduce the computational cost on time and frequency dimensions. Batch normalization and Relu activation are applied precede each convolution (omitted in the Table for simplicity).} 
  \label{RCNN-CTC-arctec}
  \vskip 0.15in
  \centering
  \begin{tabular}{ccc}
    \hline
    Layers & [filter, $\#$map $\times$ width] & Stride \\ \hline
    conv1 & [41 $\times$ 11, 32 $\times$ 1] & (2,2)\\
    \convname{ResBlock1} & \blocka{64$\times$2}{N} & (1,1) \\[\cellheight]
    \convname{ResBlock2} & \blocka{128$\times$2}{N} & (1,1) \\[\cellheight]
    \convname{ResBlock3} & \blocka{256$\times$2}{N} & (2,1) \\[\cellheight]
    \convname{ResBlock4} & \blocka{512$\times$2}{N} & (2,2) \\
    Fully-Connected & - & - \\
    CTC & - & -\\
    \hline
  \end{tabular}
  \vskip -0.1in
\end{table}

In general, convolutions require a context window, thus $conv1$ is set by considering the input feature dimension and the empirical window size. We also employ batch normalization (BN) \cite{ioffe2015batch} technique in RCNN-CTC, which is used for normalizing each layer’s input to reduce internal covariance shift. BN speeds up training and acts as a regularizer. The standard formulation of BN for CNN can be readily applied here, and we do not need the sequence-wise normalization of RNN \cite{amodei2015deep}. Moreover, strided convolutions are an essential element of CNN. For RCNN-CTC applying striding is also a natural way to reduce the computational cost on time and frequency dimensions. We find that RCNN-CTC's performance is sensitive to the stride on the time dimension but not on the frequency dimension. Unlike ResNets used in computer vision where $ResBlock2$ and $ResBlock3$ need to 
be set deeper than $ResBlock1$ and $ResBlock4$ to describe the shape or skeleton, each ResBlock has almost the same importance in ASR (\textit{i.e.}, $N$ is identical for each ResBlock). 
In summary, our proposed RCNN-CTC has a deeper and wider network architecture, compared to the existing CNN-based systems in ASR.

\subsection{CTC}
Traditional acoustic model training is based on frame-level labels with cross-entropy criterion (CE), which requires a tedious label alignment procedure. Following \cite{hannun2014deep,amodei2015deep,miao2015eesen}, we adopt the CTC objective \cite{graves2006connectionist} to automatically learn the alignments between speech frames and their label sequences, leading to an end-to-end training. 

To align the network outputs with the label sequences, an intermediate representation of CTC path is introduced in \cite{graves2006connectionist}. The label sequence $\textbf{z}$ can then be mapped to its corresponding CTC paths. It is a one-to-many mapping because multiple CTC paths can correspond to the same label sequence. For example, both ``A A $\phi$ $\phi$ B C $\phi$" and ``$\phi$ A A B $\phi$ C C" are mapped to label sequence ``A B C", where $\phi$ is the blank symbol. We denote the set of CTC paths for $\textbf{z}$ as $\Phi(\textbf{z})$. The likelihood of \textbf{z} can thus be evaluated as a sum of the probabilities of its CTC paths:
\begin{equation}
    P(\textbf{z}|\textbf{X}) = \sum_{\textbf{p} \in \Phi(\textbf{z})} P(\textbf{p}|\textbf{X}),
\end{equation}
where $\textbf{X}$ is the utterance consisting of speech frames and \textbf{p} is a CTC path. Given this distribution, we can derive the objective function of sequence labeling $\ln P(\textbf{z}|\textbf{X}) $. Since the objective function is differentiable, we can back-propagate these errors and further update the network parameters.

\section{CTC-based System Combination}
With regard to the conventional system combination, its performance improvement is little, due to the slight difference among subsystems. Therefore, we propose a system combination method which takes the diversity and complementary among subsystems into account. As a result, our proposed system combination can obtain an absolute WER reduction of $1\%$ on WSJ and Tencent Chat data sets. 


\subsection{Subsystems Selection}
Our selection of subsystems is guided by the following principles: compared to the transcription (ground truth) $G$, we first figure out the correct part/words $C_i$ in the decoding text of each subsystem $i$. We then search for the combination of subsystems and compute their union set of correct words $U = \bigcup_i C_i$. We define a maximal correct word rate (MCWR) as the selection criterion:
\begin{equation}
MCWR = \sum_{w \in G} \frac{\mathbb{I}(w \in U)}{|G|},
\end{equation}
where $\mathbb{I}(\cdot)$ is the indicator function that takes 1 if $(\cdot)$ is true and 0 otherwise, and $|G|$ is the length of ground truth $G$. 

\begin{figure}[!t]
\vskip 0.2in
\centering
\includegraphics[scale=0.6]{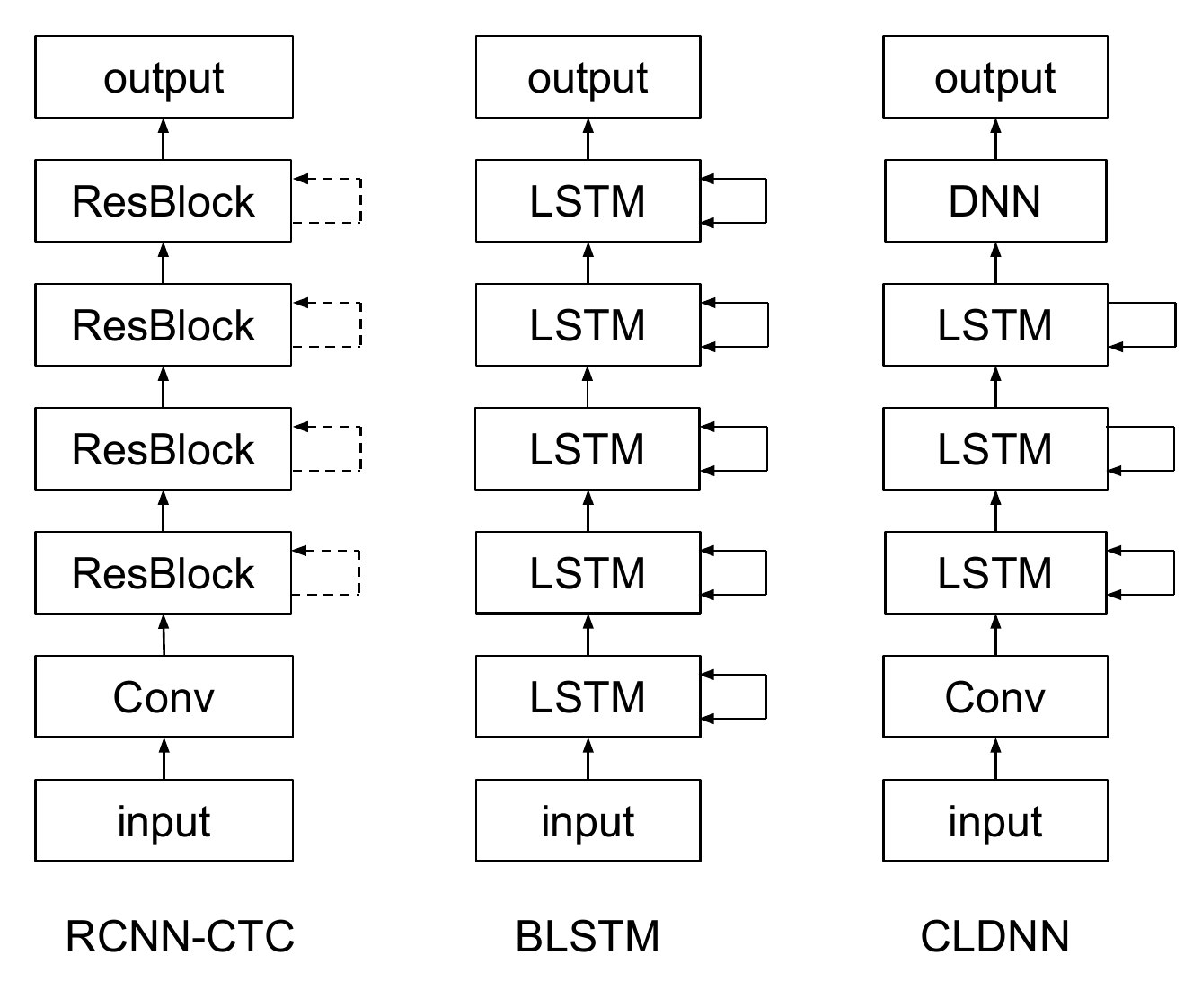}
\caption{The network architectures of RCNN-CTC, BLSTM and CLDNN. The dashed lines in RCNN-CTC are the residual connections. The bidirection arrows at the right side of LSTM in BLSTM and CLDNN represent the bidirection LSTM layer. }
\label{network}
\end{figure}

Our goal is to select the combination which achieves the highest MCWR while using a minimal number of subsystems at the same time. Through this method, we can use the least cost to find subsystems which are mutually complementary. This also provides a guideline to choose the combination which has a balance between recognition accuracy and combination costs. In our experiments, a small held-out data of WSJ is used to search for an optimal system combination via MCWR metric. Therefore, the following three subsystems\footnote{On the held-out data, two subsystems cannot achieve an acceptable MCWR (0.95) but three subsystems have already obtained a very high MCWR (0.98), while four subsystems' MCWR (0.98) is almost the same to the three ones.} are selected: 1) The proposed RCNN-CTC in Section 3; 2) BLSTM \cite{sak2014long} which consists of several bidirectional LSTM layers; and 3) CLDNN \cite{sainath2015learning}, which consists of convolutional layers, LSTM layers and DNN layers. Figure \ref{network} demonstrates these subsystems network architectures. Due to the MCWR metric and their heterogeneous structures, they may be mutually complementary to each other, which is confirmed in our experiments.

An illustrative example in WSJ data set is given below to explain the complementary of subsystems. Here, the ground truth is: \\
{\color{red} CONTACTS STILL INSIDE OWENS CORNING HELP TOO} \\
and the output sentences of the three subsystems are: 
\begin{itemize}
\item[1.] CONTACTS STILL INSIDE \underline{OWNS} CORNING \underline{HELPED} TOO 
\item[2.] \underline{CONTACT} STILL INSIDE OWENS CORNING \underline{HELPED} $\rule{0.8cm}{0.15mm}$
\item[3.] \underline{CONTACT} STILL INSIDE \underline{OWNS} CORNING HELP TOO
\end{itemize}

The incorrect words in the output of each subsystem are marked underline. We can see that each subsystem has its own defect (\textit{i.e.}, incorrect words). But the incorrect words are different among the three subsystems, and they can be mutually corrected to a certain extent. For example, compared to the ground truth word ``CONTACTS'', the word ``CONTACT'' is incorrect in subsystem 2 and 3 while it is correct in subsystem 1. The hope is that via system combination, we can leverage multiple systems to have more correct words. 

In summary, we select three different types of subsystems for combination, including CNN-based subsystems (RCNN-CTC), LSTM-based subsystems (BLSTM) and their mixture (CLDNN). We argue that CNN can have a global view on a long utterance via hierarchical feature abstraction from bottom up, while LSTM can capture the sequence information contained in long sentences. The system combination can thus realize both advantages.

\subsection{Challenges}
\begin{figure}[!t]
\vskip 0.2in
\centering
\includegraphics[scale=0.55]{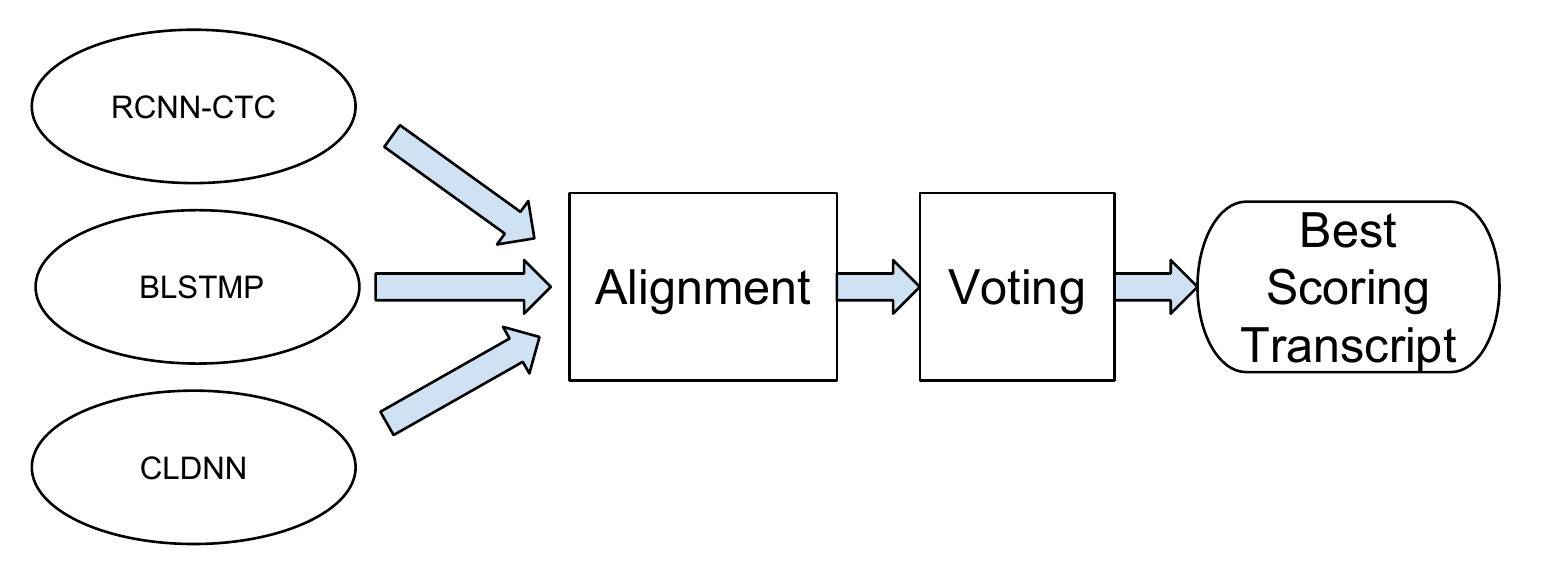}
\caption{The procedure of the proposed CTC-based system combination. Alignment including time normalization and WTN construction is applied to the outputs of each subsystem, followed by a voting scheme employed to produce the best scoring transcript.}
\label{combine}
\end{figure}

Our proposed single system RCNN-CTC uses a CTC output. The system combination is thus CTC based and different from the frame-wise CE-based one, since the peak responses of CTC in each subsystem may mismatch. Besides, the output likelihood of each subsystem is not at the same scale of time, which confuses the decoding process of the Weighted Finite-State Transducer (WFST) used in our experiments. Inspired by ROVER \cite{fiscus1997post}, we propose our CTC-based system combination method (Figure \ref{combine}) as follows. For each subsystem, after decoding with the WFST graph (TLG), 1-best hypothesis\footnote{We have tested top $N$ ($N>1$) hypotheses and found that the results are no better, see Section 5.4 for details.} with confidence score is prepared for the following processes. Alignment and composition are applied to the hypotheses of various subsystems to generate a single composite word transition network (WTN). Once the WTN is generated, we select the best scoring word from each branching path by a voting scheme to produce a new hypothesis. 

\subsection{Alignment}
\textbf{Time Normalization.} 
With regard to each subsystem, after searching the best lattice path, we get a hypothesis sequence with each item involving a label, confidence score, starting time and duration time, which may not be at the same scale due to the CTC decoding. Therefore, we need to unify the time length and rescale starting/duration time to the same scale before constructing a WTN. 

\begin{figure}[!htbp]
\vskip 0.1in
\centering
\includegraphics[scale=0.5]{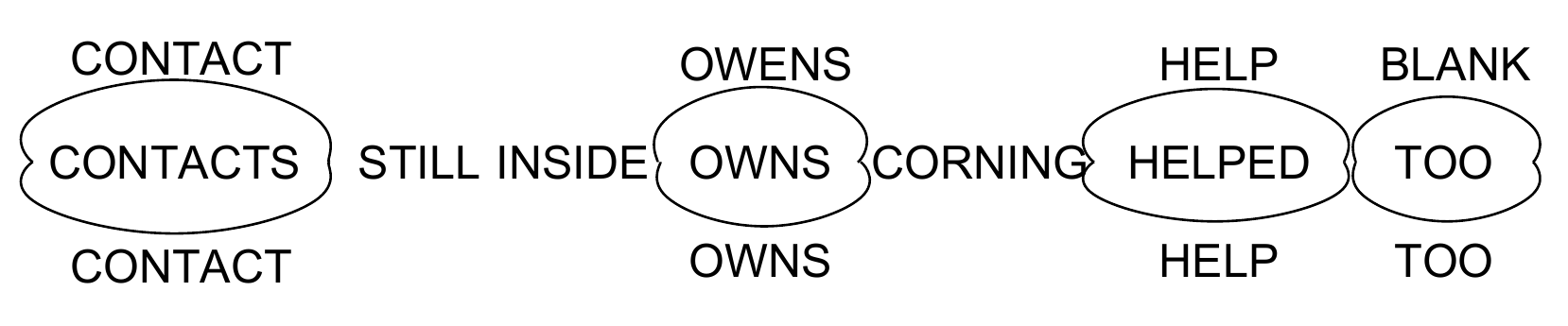}
\caption{Final single and composite WTN for the example in section 4.1. For simplicity, we omit the branching point for correct words in all three subsystems, \textit{e.g.}, STILL. }
\label{WTN}
\vskip -0.1in
\end{figure}

\textbf{WTN Construction.}
After time normalization, we can align and combine the hypotheses sequences into a single composite WTN. In particular, one of the sequences is chosen as the base WTN (WTN-BASE), and other sequences are added to WTN-BASE word by word. Comparing the word in the sequence and the corresponding word in WTN-BASE, we adopt different operations for different conditions. 1) Correction. A branching point is created and the word transition arc is added to WTN-BASE; 2) Substitution. A branching point is created and the word transition arc is added to WTN-BASE; 3) Deletion. A branching point is created and the BLANK transition arc is added to WTN-BASE; 4) Insertion. A sub-WTN is created and inserted between the adjacent nodes in WTN-BASE to record the fact. Following the above procedure, we iteratively combine the lattice words until the final composite WTN is generated. Considering the example in section 4.1, if we select the output of subsystem 1 as WTN-BASE, the first word in WTN-BASE is ``CONTACTS'' while it is ``CONTACT'' in subsystem 2 and 3. This satisfies the substitution condition, we thus create a branching point and add the word transition arc of ``CONTACT'' to WTN-BASE. The rest words are processed in a similar way until the final single composite WTN is generated in Figure \ref{WTN}. 


\subsection{Voting}
Once the composite WTN has been generated, a voting module is employed to select the best scoring word sequence by searching the WTN. According to ROVER, there are three voting schemes, \textit{i.e.}, voting by 1) frequency of occurrence, 2) frequency of occurrence and average word confidence, and 3) frequency of occurrence and maximum confidence. Generally, the third voting scheme, \textit{i.e.}, frequency of occurrence and maximum confidence, usually reports the best results \cite{fiscus1997post}, which is thus adopted in our system combination. As for the choice of confidence score, we use the minimum Bayes risk score \cite{xu2011minimum} to serve as maximum confidence. 






\section{Experiments}
We analyze the performance of our proposed RCNN-CTC and CTC-based system combination on a benchmark data set, Wall Street Journal (WSJ), and a large mobile chat data set, Tencent Chat from Tencent company. Tencent Chat data set contains about 2.3 million utterances which account for 1400 hours speech data. 

\subsection{Experimental Setup}
For WSJ data set, we use the standard configuration si284 for training, eval92 for validation and dev93 for test. Our input features are 40 dimensional filterbank features with delta and delta-delta configuration. The features are normalized via mean subtraction and variance normalization on the speaker basis. 

For Tencent Chat data set, we use about 1400 hours internal speech data for training and an independent 2000 utterances for test. Our input features are 40 dimensional filterbank combined with 3 dimensional pitch features, and are normalized by per utterance mean and variance as there is no speaker information.

We use the Kaldi recipe \cite{povey2011kaldi} to prepare the dictionary for WSJ and Tencent Chat data sets. It in fact uses CMU dictionary and Sequitur G2P to prepare phone sequences for both English and Chinese words. Finally, we have 118 phones served as acoustic model output labels. 

Our decoding follows the WFST-based approach in EESEN \cite{miao2015eesen}. As for the language model, we apply the WSJ pruned trigram language model with expanded lexicon \cite{povey2011kaldi} in the ARPA format on WSJ data set. For Tencent Chat data set, we use 5-gram language model trained with about 6 billion tokens (120K vocabulary) corpus from an internal data set. All the networks use phone-based training by stochastic gradient descent optimization (SGD). The learning rates are initialized to be in the range of $4\times10^{-5}$ to $1\times10^{-4}$, and are exponentially decayed by a factor of 0.1 after every 10 epochs during training.




\subsection{Results on WSJ data set}
We compare our proposed single system RCNN-CTC with several commonly used neural network baseline systems in ASR, \textit{i.e.}, BLSTM \cite{sak2014long}, CLDNN \cite{sainath2015learning} and VGG \cite{simonyan2014very}. 

BLSTM is implemented according to \cite{miao2015eesen}, which uses 4 bidirectional LSTM layers. At each layer, both the forward and the backward layers comprise 320 hidden units. CLDNN is implemented following \cite{amodei2015deep}, which contains 3 convolutional layers, 3 bidirectional LSTM layers and 2 fully-connected layers. The kernel sizes of the three convolutional layers are (11, 21), (11, 11), (3, 3), and the strides are (3,2), (1,2), (1,1) respectively. Batch normalization and Relu activation function are also employed. Each LSTM layer consists of 896 hidden units and 2 fully-connected layers have 896 and 74 units respectively. VGG is implemented according to \cite{yu2016deep}, which has 14 layers. First, there are 3 convolutional layers with small $3\times3$ filters and $96$ feature maps, followed by a max-pooling layer. Then, 4 convolutional layers with $192$ feature maps and 4 convolutional layers with $384$ feature maps are added, all using $3\times3$ filters and max-pooling at the end. With regard to RCNN-CTC, we adopt the parameters in Table \ref{RCNN-CTC-arctec} with $N = 2$. 

\begin{table}[!t]
    \centering
    \caption{WER ($\%$) of single systems on WSJ data set. The lowest WERs are in boldface. RCNN-CTC achieves the lowest WER on both eval92 and dev93, compared to other single systems. The third column shows the relative WER reductions (WERR ($\%$)) of RCNN-CTC compared to other systems. The last two rows of BLSTM show that CTC can obtain lower WER than CE.}
    \label{single_wsj}
    \vskip 0.15in
    \begin{tabular}{lcc}
    \hline
    Single & WER & WERR \\ 
    System & (eval92/dev93) & (eval92/dev93) \\ \hline
    RCNN-CTC & \textbf{5.35/8.99} & -/- \\ 
    VGG+CTC & 5.39/9.05 & 0.74/0.66 \\ 
    CLDNN+CTC & 5.39/9.02 & 0.74/0.33 \\ 
    BLSTM+CTC & 5.48/9.12 & 2.37/1.43 \\ 
    BLSTM+CE & 5.54/9.36 & 3.43/3.95 \\ \hline    
    \end{tabular}
    \vskip 0.1in
\end{table}

\begin{table*}[!t]
    \centering
    \caption{WER ($\%$) of combined systems on WSJ data set. The lowest WERs are in boldface. WERR ($\%$) is the relative WER reduction of each combination compared to the best single system RCNN-CTC.  (For simplicity, we write RCNN in palce of RCNN-CTC here.)}
    \label{combine_wsj}
    \vskip 0.15in
    \begin{tabular}{lcc}
    \hline
    Combined & WER & WERR\\ 
    System & (eval92/dev93) &  (eval92/dev93) \\ \hline
    BLSTM+VGG+CLDNN & 4.62/7.93 & 13.64/11.79 \\ 
    RCNN+VGG+CLDNN & 4.70/8.04 & 12.15/10.57 \\  
    RCNN+BLSTM+VGG & 4.59/7.87 & 14.20/12.46 \\ 
    RCNN+BLSTM+CLDNN (Ours) & \textbf{4.29/7.65}  & \textbf{19.81/14.91} \\  
    \hline    
    \end{tabular}
    \vskip 0.1in
\end{table*}

Table \ref{single_wsj} compares our proposed single system RCNN-CTC with baseline systems on WSJ data set. For all systems trained with CTC loss, we can observe that RCNN-CTC obtains WER of $5.35\%$ and $8.99\%$ on eval92 and dev93 respectively\footnote{Lower WER results on WSJ data set were reported in Kaldi Speech Recognition project\footnotemark, however, these results were achieved using additional techniques including speaker-adaptive features, splice context for data preparation, and iVector for instantaneous adaptation.} \footnotetext{https://github.com/kaldi-asr/kaldi}, which slightly outperforms BLSTM, VGG and CLDNN. We speculate the slight gain may be limited to the small data size of WSJ, as the proposed RCNN-CTC cannot demonstrate its full system strength. We will observe much larger gain of RCNN-CTC \textit{vs.} other systems when a larger Chat data set is used in Section 5.3. Moreover, we show additional results in Table \ref{single_wsj} where we compare systems trained with CTC and CE. Here, we only take BLSTM system as an example. The results are similar for other systems. For BLSTM+CE, we use GMM-HMM system \cite{rabiner1989tutorial} to generate the label alignment to train. The GMM-HMM system is trained with the maximum likelihood (ML) criterion and refined with the boosted maximum-mutual-information (BMMI) sequence-discriminative training criterion. As can be seen from the last two rows of Table \ref{single_wsj}, BLSTM+CTC slightly outperforms BLSTM+CE, whereas the former can be trained in an end-to-end manner while the latter requires label alignment.

We next proceed with the system combination experiments on WSJ data set. For a fair comparison, we only consider combinations of three subsystems\footnote{As mentioned in Section 4.1, on a held-out data set, two subsystems cannot achieve an acceptable MCWR (0.95) but three subsystems have already obtained a very high MCWR (0.98), while four subsystems' MCWR (0.98) is almost the same to the three ones.} among the four: RCNN-CTC, VGG-CTC, CLDNN-CTC and BLSTM-CTC. Table \ref{combine_wsj} shows all four possible combinations and their WER on eval92 and dev93 respectively. It is worth pointing out that the WERs of subsystem may not be a useful metric for selecting subsystems for combination. Instead, it is the complementary among subsystems that really matters. For example, RCNN-CTC, VGG and CLDNN are top 3 single systems with regard to WER in Table \ref{single_wsj}, while their system combination results are $4.70\%/8.04\%$ on eval92 and dev93 respectively, which is the worst in Table \ref{combine_wsj}. While our system combination has the lowest WER of $4.29\%/7.65\%$, which indicate the effectiveness of MCWR subsystem selection method. The fact that both top 2 system combinations including RCNN-CTC suggests its supremacy over other systems. Moreover, we notice that the WERs of combined systems are all lower than the single system results in Table \ref{single_wsj}, indicating that system combination can always boost the recognition accuracy. Note that our proposed system combination achieves an absolute WER drop of $1.06\%$ and $1.34\%$ (or relative drop of $19.81\%$ and $14.91\%$) on eval92 and dev93 respectively compared to the best single system RCNN-CTC. 





\subsection{Results on Tencent Chat data set}
In the following, we explore the performance of RCNN-CTC and system combination on a large Chat data set. Here, we only demonstrate the results trained with CTC loss to avoid tedious label alignment work in CE. Baseline systems are the same to those in Section 5.2, but some network parameters are slightly adjusted. CLDNN uses the same network architecture, but the kernel sizes of the three convolutional layers are (11, 11), (5, 5), (3, 3), and the strides are (3,1), (1,1), (1,1) respectively. As for RCNN-CTC, we again adopt the parameters in Table \ref{RCNN-CTC-arctec}, where the difference is $N = 5$. With regard to BLSTM and VGG, parameters of these systems are the same as in Section 5.2.

\begin{table}[!t]
    \centering
    \caption{WER ($\%$) of single systems on Tencent Chat data set. The lowest WER is in boldface. The WER differences between RCNN-CTC and VGG, CLDNN, BLSTM are significant. WERR ($\%$) is the relative WER reduction of RCNN-CTC compared to other systems.}
    \label{single_wechat}
    \vskip 0.15in
    \begin{tabular}{lcc}
    \hline
    Single System & WER & WERR\\ \hline
    RCNN-CTC & \textbf{14.26} & - \\ 
    VGG+CTC & 15.03 & 5.12 \\
    CLDNN+CTC & 14.94 & 4.55 \\ 
    BLSTM+CTC & 15.03 & 5.12 \\ 
    \hline    
    \end{tabular}
\end{table}

\begin{table}[!t]
    \centering
    \caption{WER ($\%$) of combined systems on Tencent Chat data set. The lowest WER is in boldface. WERR ($\%$) is the relative WER reduction of each combination compared to the best single system RCNN-CTC. (For simplicity, we write RCNN in palce of RCNN-CTC here.)}
    \label{combine_wechat}
    \vskip 0.15in
    \begin{tabular}{lcc}
    \hline
    Combined System & WER & WERR \\ \hline
    BLSTM+VGG+CLDNN & 13.58 & 4.77 \\ 
    RCNN+VGG+CLDNN & 13.41 & 5.96 \\  
    RCNN+BLSTM+VGG & 14.01 & 1.75\\ 
    RCNN+BLSTM+CLDNN (Ours) & \textbf{13.33} & \textbf{6.52} \\ 
    \hline    
    \end{tabular}
\end{table}

Table \ref{single_wechat} summarizes the WERs of single systems on Tencent Chat data set. Compared to VGG, CLDNN and BLSTM, RCNN-CTC performs the best and obtains an absolute WER reduction of $0.77\%$, $0.68\%$ and $0.77\%$, or relative WER reduction of $5.12\%$, $4.55\%$ and $5.12\%$ respectively. Furthermore, Table \ref{single_wechat} confirms the advantages of deep CNN architecture for ASR tasks on large data sets. RCNN-CTC and VGG are both CNN-based systems, while RCNN-CTC has residual connections as described in Section 3, which allow it to have very deep network depth (RCNN-CTC 40 layers \textit{vs.} VGG 14 layers) and thus achieve higher accuracy. 

Similar to the experiments on WSJ data set, we also carry out a series of experiments on Tencent Chat data set to further assess the proposed system combination. We again consider combinations of three subsystems only, with WER of all combinations are collectively listed in Table \ref{combine_wechat}. Similar to the WSJ system combination, the combination of RCNN+BLSTM+CLDNN outperforms others, due to the maximal complementary of these three subsystems described by MCWR. As can be noticed, top 2 system combinations also both choose RCNN-CTC as one base subsystem, which reveals its superb capacity in ASR. WERR is the relative WER reduction of each combination with respect to the best single system RCNN-CTC in Table \ref{single_wechat}. Our proposed system combination can achieve WER of $13.33\%$, which accounts for an absolute WER drop of $0.93\%$ or relative drop of $6.52\%$ compared to RCNN-CTC.

In summary, the experimental results are representative to reveal the effectiveness of our proposed single system RCNN-CTC and CTC-based system combination.





\subsection{Analysis and Discussion}

\textbf{Choice of 1-best \textit{vs.} N-best in system combination.}
As stated in Section 4, we choose 1-best hypothesis for combination, because we find that N-best is no better than 1-best in our experiments, as shown in Table \ref{1best}. Here N-best (N=10) distinct hypotheses of each subsystem are prepared for combination. Firstly, if we use the voting scheme in Section 4.4, \textit{i.e.}, maximal confidence score voting, choosing N-best does not offer any further benefits. Although N-best hypotheses make the WTN contain more branchings and words choices, maximal confidence score voting almost gets the same result as with 1-best hypothesis. The first two rows of Table \ref{1best} verify the above conclusions. Moreover, we conduct another experiment using frequency of occurrence as voting score for N-best subsystems combination. We find that the results are close to 1-best on WSJ data set but slightly worse on Tencet Chat data set. This is because that one subsystem's error may repeat many times in N-best hypotheses, which distorts the following frequency-based voting. Furthermore, considering the computational cost of N-best hypotheses, 1-best from each subsystem with maximal confidence score may be preferred. 

\begin{table}[!htbp]
\centering
\caption{The comparisons of choosing 1-best and N-best hypotheses of each subsystem for system combination. The performance of N-best hypotheses is no better than 1-best regardless which voting schemes are used.}
\label{1best}
\vskip 0.15in
\begin{tabular}{ccc}
\hline
 & WER-WSJ($\%$) & WER-Chat($\%$) \\\hline
1-best (Ours) & 4.29/7.65 & 13.33\\
N-best (Ours) & 4.30/7.65 & 13.32 \\
N-best (Frequency) & 4.32/7.76 & 13.67\\
\hline
\end{tabular}
\end{table}

\section{Conclusions}
In this paper, we proposed a novel residual convolutional neural networks architecture trained with CTC loss (RCNN-CTC) for ASR. We argued that CNN is suited to exploit local correlations of human speech signals in both time and frequency dimensions, and has the capacity to exploit translational invariance in signals. In our proposed RCNN-CTC, we employ a wide and deep CNN architecture (more than 40 layers) with residual connections, which owns more expressive power and better generalization capacity. RCNN-CTC can be trained in an end-to-end manner thanks to the adoption of CTC loss, which effectively avoids the tedious frame alignment process. Furthermore, we proposed a CTC-based system combination via subsystems selection, alignment and voting procedures. Experiments on WSJ and Tencent Chat data sets show that, among widely used neural network systems in ASR, RCNN-CTC obtains the lowest WER. In addition, significant WER reductions are further obtained via our proposed system combination. For example compared to RCNN-CTC, the proposed system combination further results in relative WER reductions of $14.9\%$ and $6.52\%$ on WSJ dev93 and Tencent Chat data sets respectively.


\bibliography{ref}
\bibliographystyle{icml2016}

\end{document}